\documentclass[10pt,twocolumn,letterpaper]{article}

\usepackage[pagenumbers]{cvpr} 

\usepackage{graphicx}
\usepackage{amsmath}
\usepackage{amssymb}
\usepackage{booktabs}
\usepackage{multirow}
\usepackage{soul}

\usepackage[pagebackref,breaklinks,colorlinks]{hyperref}

\usepackage[capitalize]{cleveref}
\crefname{section}{Sec.}{Secs.}
\Crefname{section}{Section}{Sections}
\Crefname{table}{Table}{Tables}
\crefname{table}{Tab.}{Tabs.}

\def\cvprPaperID{*****} 
\def\confName{CVPR}
\def\confYear{2022}

\begin{document}

\title{Modality Agnostic Efficient Long Range Encoder}
\author{
Toufiq Parag \\
Amazon Prime Video \\
\texttt{\small toufiq.parag@gmail.com}
\and
Ahmed Elgammal \\
Rutgers University \\
\texttt{\small elgammal@cs.rutgers.edu}
}



\maketitle

\begin{abstract}

The long-context capability of recent large transformer models can be surmised to rely on techniques such as attention/model parallelism, as well as hardware-level optimizations. While these strategies allow input lengths to scale to millions of tokens, they do not fundamentally mitigate the quadratic computational and memory complexity of the core attention mechanism. In this paper, we address the challenge of long-context processing on a single device using generic implementations by reducing the quadratic memory footprint and inference cost. Existing approaches to extend the context length for  generic single device implementations  -- such as token merging and modified attentions -- are often modality specific and attain a suboptimal tradeoff between accuracy and efficiency. To overcome these limitations, we propose MAELRE (Modality Agnostic Efficient Long Range Encoder), a unified and efficient transformer architecture designed for long-range encoding across diverse modalities. MAELRE integrates token merging with attention approximation, progressively merging tokens at different stages of internal computational blocks. It employs a lightweight attention approximation when the number of tokens is large, and switches to standard dot-product attention as the sequence becomes shorter through successive aggregation. We demonstrate that MAELRE achieves superior accuracy while reducing computational cost compared to existing long-context models on classification tasks spanning multiple modalities, including text, time series, audio, and vision.

\end{abstract}

\section{Introduction}
\label{S:INTRO}


Recent large language models (LLMs) and vision language models (VLMs) can operate on several hundred thousand to millions of tokens. Although the underlying technology has not been published, one can speculate that these models employ some combinations parallelized attention or model~\cite{liu2023ring} and hardware level optimizations~\cite{dao2022flashattention} on a large cluster of relatively expensive devices.  These approaches serve as workarounds for the $O(N^2 + N D)$ memory and $O(N^2 D)$ computational complexities imposed by the dot-product attention mechanism, where $N$ denotes the number of input tokens and $D$ their dimensionality. However, parallelization or device-specific optimizations do not fundamentally reduce the quadratic memory and compute costs of a generic implementation of core attention. Instead, they typically require arrays of premium GPUs with different memory architectures tailored for hardware optimizations. In this paper, we focus on the problem of long-context processing in generic (i.e., device-agnostic) implementations of the attention mechanism which will not only enable single, low-cost GPUs to handle long inputs effectively, but also can potentially amplify the benefits of the aforementioned methods when used in combination. Specifically, we address the challenge of \emph{encoding} long-context inputs across diverse modalities using a single transformer model deployed on a single device, without relying on device-specific optimizations.

Many efforts have been made to enable transformer models to process long contexts. Relevant studies, that focus on single model and device architecture (excluding hardware level implementations, e.g.,~\cite{dao2022flashattention,rabe2021selfattention}), appear to explore two avenues in this regard. One of the directions is to progressively reduce the number of tokens $N$ by aggregation at different stages of the computation within the model. This technique is popular in visual domain where rudimentary, and often superfluous, components of target concepts are gradually coalesced to composite representations relevant to output tasks in deeper layers~\cite{liu2021swin, yu2022metaformer,li2022efficientformer, fan2021multiscale}. The other strategy adjusts the attention mechanism to mitigate $O(N^2)$ cost by restricting pairs of tokens to attend to~\cite{beltagy2020longformer, zaheer2020bigbird} or through efficient kernel approximations~\cite{choromanski2021rethinking, katharopoulos20rnn}. Although these techniques were instrumental in breaking the barrier of limited input length and have sound theoretical foundations, in practice, they have not been able to consistently supersede or match the vanilla transformer accuracy on different tasks~\cite{dao2022flashattention, phang2023investigating, zaheer2020bigbird,tay2021long}.

Conceptually, it is rational to speculate that merging elementary information could also be advantageous for long context processing for different modalities beyond vision. Several recent studies illustrate how language transformers indeed consolidate the primitive facts to establish relationship in the initial layers of computation~\cite{geva2023dissecting, wang2022interpretability}. Because primitive facts are intrinsically aggregated, it is perhaps redundant to represent every token in all layers of the architecture.  Integrating token aggregation with the aforementioned efficient attention algorithms should further extend the input size a model can handle. It becomes crucial to determine a suitable attention approximation technique, and apply it intelligently in tandem with token merging, for such a combination to achieve high accuracy while maintaining memory and computational efficiency.
\begin{figure*}[t]
\vspace{-0.12in}
\begin{center}
\centerline{\includegraphics[width=0.825\textwidth, height=3.5in]{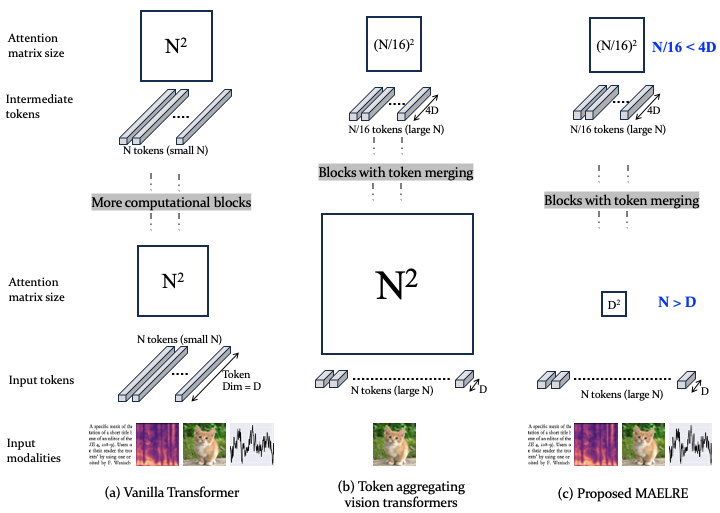}}
\vspace{-0.2cm}
\captionsetup{font={scriptsize,stretch=0.9}} 
\caption{Illustrating advantage of the proposed MAELRE through attention matrix size with respect to token number and dimensions. The number and dimensions of tokens, and consequently, the size of attention matrix remains same across all computational blocks of vanilla transformer (left). For many vision transformers, the attention matrix grows very large at the initial layers due to large number of tokens but reduces in deeper layers due to token aggregation (middle). In the proposed MAELRE design, we apply an approximate $O(D^2)$ attention when number of tokens $N$ is larger than token dimensions $D$ in the early layers (right). In the deeper layers, when $N<D$ due to token merging, the standard dot product attention is utilized. The proposed MAELRE is a general design that can be applied different modalities, e.g., text, audio, time series, vision.}
\end{center}
\label{F:INTRO}
\vspace{-0.8cm}
\end{figure*}

In this paper, we propose an efficient transformer encoder that can process long input of multiple modalities. The proposed modality agnostic encoder (MAELRE) combines the token reduction and attention approximation mechanism in a non-trivial fashion to maximize the  benefits.  Our design reduces the number of tokens at different depths of the model. The tokens in the early shallow layers are represented by fewer number of features than the those representing  more informative composite tokens in deeper layers. We employ a kernel based $O(D^2)$ approximation and the original $O(N^2)$ dot product attention in the shallow and deep layers respectively to minimize memory and computation costs. In particular, at shallow layers, where $N>D$, we utilize a modified attention with $O(D^2)$ memory requirement. In deeper layers, where $N<D$ due to token aggregation, we use the $O(N^2)$ dot product attention. Essentially, our design incurs a memory cost of $O(M^2)$ where $M = \min(N,D)$ which leads to substantial cost reduction when $N$ is large. Figure 1 illustrates the advantage of proposed approach. 

We empirically justify (Section~\ref{S:RES_MESH}) our hypothesis that harnessing preliminary information from all tokens using approximate attention captures more knowledge than calculating dot product attention on a subset and ignoring a significant number of tokens as suggested in~\cite{beltagy2020longformer, zaheer2020bigbird, child2019generating}. Furthermore, we argue and experimentally validate (Section~\ref{S:ABL_ATTN}) that applying attention approximation in the early layers, which operates on basic facts encoded in a large number of tokens, does not compromise accuracy for long range problems within the scope of this paper. 

The contribution of this study can be summarized as follows. 1) We present an efficient transformer architecture MAELRE for processing long context on a single device through an innovative combination of token aggregation and attention approximation techniques. The proposed strategy employs an $O(D^2)$ attention when $N$ is large and an $O(N^2)$ attention when $N$is small, significantly curtailing memory and computational costs due to the progressive merging of tokens while maintaining accuracy. 2) We demonstrate that the proposed encoder is generally applicable to multiple modalities, including text, time series, audio and image; and is more accurate and less costly than existing long range models. 

Empirically, MAELRE achieves $1\%$ higher accuracy in a text classification task while incurring only $3\%$ and $1\%$ memory and computational costs, respectively, than the next best model. The proposed architecture can fit 24 × 17000 text tokens on a single low-cost device (16GB T4 GPU) without any hardware optimization. Furthermore, it achieves state of the art accuracy on VGGSound audio-only classification (57.8\%) and Mosquito sound time series dataset (95.42\%) while requiring roughly
half the memory and FLOPs of the closest competitor. Employing multi-device or model approaches~\cite{liu2023ring,bulatov2023scaling,li-etal-2023-sequence,yu2023megabyte} on proposed MAELRE architecture is expected to further enhance the benefit of our model for many applications.



\section{Related Works}
\label{S:RELATED}

The machine learning literature has been enriched by attention approximation techniques to expand the input length the transformer architecture can handle efficiently. One prominent approach is to restrict the pairs of tokens to compute attention. Studies like~\cite{child2019generating, beltagy2020longformer, zaheer2020bigbird,guo2022longt5} specify a restricted set of token pairs for attention, some with mechanisms to capture part of the global context~\cite{zaheer2020bigbird,guo2022longt5}. The other direction is to approximate the dot product attention using kernel feature decomposition~\cite{choromanski2021rethinking, katharopoulos20rnn}. Yet another approach that gained attention was to consolidate the query tokens either by linear projection~\cite{wang2020linformer} or hashing~\cite{kitaev2020reformer} among other techniques. We hypothesize that using an approximate attention with lower cost among all tokens is more effective than applying original dot product attention on a subset of tokens~\cite{child2019generating, beltagy2020longformer, zaheer2020bigbird,guo2022longt5} for processing long range context, especially in the early layers where the basic information are consolidated. In addition, we demonstrate an efficient and effective utilization of a combination of approximate and dot product attentions for long range inputs of multiple modalities.

In the vision literature, token reduction in conjunction with expansion of feature resolution and attention head has become a common practice in recent years~\cite{liu2021swin, graham2021levit, heo2021rethinking, bolya2022token, yu2022metaformer, fan2021multiscale, li2022mvitv2, Pan_2021_ICCV}. In these studies, different approaches are used for feature reduction. For example, ~\cite{liu2021swin, heo2021rethinking} reduce token through convolution, ~\cite{graham2021levit, fan2021multiscale} subsample the queries, ~\cite{li2022efficientformer,yu2022metaformer, Pan_2021_ICCV} apply pooling, ~\cite{bolya2022token, marin2023token} merge similar tokens by matching or clustering.  Similar to  ~\cite{liu2021swin, heo2021rethinking, li2022efficientformer,yu2022metaformer}, we separate token aggregation from the attention mechanism and demonstrate that a convolution based merging works well for multiple modalities. It is worth distinguishing our work from the post training optimization during inference proposed for vision tasks~\cite{yin2022avit,zhao2022adavit} -- we present an efficient long range encoder for training and inference in multiple modalities.

Some recent findings suggest that token merging can also be beneficial for other modalities such as text. For example, ~\cite{geva2023dissecting, meng2022locating, wang2022interpretability} illustrate how primitive information pieces are aggregated through a subject enrichment process in the early computational layers and how knowledge is consolidated by a small fraction of computational units within language models. It is not irrational to speculate that many of these internal circuits of consolidation are shared, and perhaps are reused, by different tasks.

However, there are not many studies that investigate the advantage of token merging and approximate attention across modalities. Notable exceptions like~\cite{dai2020funnel, goyal2020powerbert} incorporate token merging for language processing, but this strategy has not become a mainstream trend. Recently, ~\cite{han2023flatten} have applied the attention approximation of~\cite{katharopoulos20rnn}, supplanted by additional modifications, to visual tasks.  We show that an encoder constructed exclusively with approximate attention~\cite{han2023flatten} is not optimal for different modalities (e.g., audio, text) and instead employs an insightful combination of dot product and approximate attentions. Other efficient long range methods that have been demonstrated to be effective in multiple modalities include~\cite{zhu2021longshort,tay2020sparse}.

In parallel, several studies explore implementation and hardware optimization for efficient attention computation on a single device~\cite{rabe2021selfattention, dao2022flashattention, dao2023flashattention2}. Notable among these works, Flash attention~\cite{dao2022flashattention} divides the query, key and value matrices into smaller blocks and optimizes I/O between different types of memory architecture present on high-end devices. Another prominent direction to reduce I/O, especially to circumvent loading key value caches during decoder inference, is to condense the multiple key, value vectors to a single one and compute attention with all queries~\cite{shazeer19multiquery, ainslie2023groupquery, deepseekai2024deepseekv2}. Recent years have also seen significant efforts to distribute the sequence or model across multiple devices coupled with global communication to enforce consistency~\cite{liu2023ring,bulatov2023scaling,li-etal-2023-sequence,yu2023megabyte}; some of these techniques have been used in building large and foundational models~\cite{bulatov2023scaling,xiong2024effective,stallone2024scaling}. These approaches are complementary to our proposed encoder design; most of them can be applied on top of the MAELRE single-device architecture to further extend the context length.

Finally, it is also worth distinguishing our effort from the state space models for long context processing~\cite{fu2022hungry, gu2023mamba} that have gained a substantial attention in recent years. Because the architecture used by these methods is fundamentally different from the transformer models we investigate, we exclude these studies from our experiments.

\section{Proposed MAELRE}\label{S:METHOD}

Our proposed MAELRE encoder architecture is composed of $B$ blocks, each consisting of multiple multihead self attention (MHSA) and multilayer perception (MLP) modules followed by a token reduction module. After each block of MHSA, MLPs, the token reduction module (Section~\ref{S:REDUC}) decreases the number of token by merging by a factor of $\nu$ and expands the dimensionality by 2. In contrast to~\cite{han2023flatten}, we leverage two forms of MHSAs with $O(N^2)$ and $O(D^2)$ memory complexities at different block depth; Section~\ref{S:ATTN} describes their forms in detail.

Data from any modality is first converted to $N$ tokens $x^0_{i_0}\in\mathbf{R}^D$ of dimensionality $D$ by a convolution of suitable modality dependent kernel $\kappa_0$ and stride $s_0$. Provided $\{x^0_{i_0}\}_{i_0=1}^N$ and a CLS token as input, the encoder generates embeddings for CLS and $x^L_{i_B} \in \mathbf{R}^{D_B}$ where $i_B\in [1, N_B], N_B = {\frac{N} {\nu^B}}, D_B = D \times {2^B}$ and $L$ is the total number of layers. In this work, we apply a fully connected layer to project CLS embedding to $C$ classes to demonstrate the encoders strength for classification. Theoretically, the embeddings can be utilized for other purposes such as matching or representation learning~\cite{liu2019roberta, radford2021learning}.

\subsection{Attention Modules}\label{S:ATTN}
Within each computational block $b$, there are $M_b$ layers of consecutive MHSA and MLP modules to attend and project the token $x^l$ to $x^{l+1}$  through residual connections and normalization. Unlike~\cite{wang2020linformer, fan2021multiscale, alayrac2022flamingo}, we do not decrease the number of tokens $N_b$ or the dimensionality $D_b$ within the MHSA module. However, based on $N_b$ and $D_b$ we utilize different attention techniques to improve efficiency.

\subsubsection{$N_b<D_b$: Dot Product Attention}\label{S:DOT}
At the deeper blocks, where the number of tokens $N_b$ is less than dimensionality $D_b$, they attend to each other via the standard dot product attention~\cite{vaswani2017attention}. The computation of dot product attention can be formalized as follows (with a slight abuse of the notation for clarity).
\vspace{-0.2cm}
\begin{equation}
    V_{\text{new}} = {\frac{\exp \bigl({\frac{1} {\sqrt{D_b}}} Q K^T \bigr)} {Z}}~ V \label{E:DOT}
\vspace{-0.2cm}
\end{equation} 
In this expression, $Q,K,V$ are the query, key and value matrices respectively that are computed from linear transformations of $X^l = [x^l_1, x^l_2, \dots, x^l_{N_b}]^T$ where $x^l_{i_b} \in \mathbf{R}^{D_b}$. The multiplier of $V$ in Equation~\ref{E:DOT} is a softmax operation: both $\exp$  and division are applied elementwise and $Z$ is a normalizing matrix to convert each row of $\exp \bigl({\frac {1} {\sqrt{D_b}}} Q K^T \bigr)$ into probability values. For our discussion, it is helpful to point out that the $\exp$ function ensures non-negativity for the computation of softmax operation and each column of $Z$ equals to $\exp \bigl({\frac{1}{\sqrt{D_b}}} Q K^T \bigr)~\mathbf{1}$ where $\mathbf{1}$ is a vector of all ones. Although the memory and computation costs for attention in block $b$ is $O(N^2_b + N_b D_b)$ and $O(N_{b}^2 D_b)$ respectively, $N_b$ is $\nu^b$ times smaller than the initial token length $N$.

\subsubsection{$N_b>D_b$: Approximate Attention}\label{S:PHI}
In shallow blocks, where number of tokens is significantly larger than embedding dimensionality $N_b > D_b$, we aim to leverage an attention with $O(D_b^2)$ complexity. We propose to utilize an approximation of scaled dot product kernels and claim that in the early stages of computation, an approximation of the dot product kernel is sufficient to capture the necessary relationship among primitive information to be propagated into the later stages. We also hypothesize that an approximation among all token pairs capture more knowledge than attention among subset of token pairs, especially at the shallow layers.

Based on the theory of kernel functions~\cite{hofmann2008kernel}, the exponential dot product kernel $\exp(a^T b)$ is equivalent to $\phi(a)^T \phi(b)$ given an infinite dimensional (hypothetical) feature function $\phi$.  The softmax operation in the dot product attention can then be reformulated as matrix multiplication between $\phi(Q)$ and $\phi(K) V^T$ due to associative property of matrix multiplication, as shown in~\cite{katharopoulos20rnn}.
\vspace{-0.2cm}
\begin{equation}
    V_{\text{new}} = {\frac {{\frac{1}{\tau}}~\phi(Q) \phi(K)^T} {Z'}}~ V = \phi(Q)~{ \frac{{\frac{1} {\tau}}~ \phi(K) V^T} {Z'}}  \label{E:APPROX}
\vspace{-0.2cm}
\end{equation} 
Here, $Z'$ is constructed by replicating the columns with  $\phi(Q) \phi(K)^T \mathbf{1}$ and $\tau$ is a scaling factor. As a result, the attention can be computed with a memory and computation costs of $O(D^2_b+ N_b D_b)$ and $O(D^2_b N_b)$ in a shallow block $b$. That is, this form of attention can handle large number of tokens since the costs are quadratic on $D_b$ instead of  $N_b$.

Since the feature function for exponential dot product kernel is infinite dimensional,  one needs to employ a kernel with closed form and efficient $\phi$ to avail the $O(D_b^2)$ complexity. Conversely, and perhaps more practically, one can instead identify a efficient and practical feature function, as~\cite{katharopoulos20rnn}proposed,  so that the dot product of two feature transformations can approximate the similarity kernel. One such option proposed by~\cite{katharopoulos20rnn} is a nonnegative function, e.g., $\psi(a) = 1+ \text{ELU}(a)$  where ELU is Exponential Linear Unit. However, we have empirically observed that this choice of $\psi$ is not critical for performance and other nonnegative functions may also be utilized for classification.  In particular, we apply and report the performances of the approximate attention (Equation~\ref{E:APPROX}) using the ELU and the following as feature functions.

\begin{itemize}
\vspace{-0.2cm}
\itemsep-0.2cm
    \item \textbf{ReLU}: $\psi(a) = max(0, a)$.
    \item \textbf{Softplus}: $\psi(a) = \log(1+ \exp(a))$.
\vspace{-0.2cm}
\end{itemize}

Furthermore,~\cite{katharopoulos20rnn} did not explicitly address  that the choice of scaling factor $\tau$ is pivotal for numerical stability. We observed a value of $\tau =  \sqrt{N_b}$ led to training convergence and used it for all our experiments.

\subsection{Reduction Modules}\label{S:REDUC}
After applying $M_b$ MHSA and MLP modules in each block $b$, the number of tokens $x^l_{i_b},~ i_b = 1, 2, \dots, N_b$ is reduced to $\tilde{x}^{l+1}_{i_{b+1}},~ i_{b+1} = 1, 2, \dots, N_{b+1} = N_b /\nu $ by a convolution  with stride $\nu$ and window size $2\nu+1$. For 2D signals such as image or audio, the tokens are reshaped as matrix and reduced via a 2D convolution with stride $({\frac{\nu}{2}},{\frac{\nu}{2}})$.  Convolution has been used for many deep net architectures for feature aggregation~\cite{liu2021swin, heo2021rethinking}. We have also experimentally found it to be more effective than other reduction techniques. Subsequently, the representation size of $\tilde{x}^{l+1}_{i_{b+1}} \in \mathbf{R}^{D_b}$ is increased by a factor of $2$ by a linear transformation to generate tokens $x^{l+1}_{i_{b+1}} \in \mathbf{R}^{2 D_b}$  to be fed into block $b+1$. 

\section{Experiments and Results}\label{S:EXP}
Our encoder architecture has $B=4$ blocks with two configurations for number $M_b$ of modules within: 1) small $\{ M_1= 1, M_2 = 2, M_3 =11, M_4 = 2\}$ and base $\{ M_1= 1, M_2 = 3, M_3 =16, M_4 = 3\}$. Unless otherwise specified, all experiments were performed with these two configurations. Given input data from any modality, the input $x^0$ to the encoder is computed using a convolution to map to $D=96$ dimensional embeddings. The stride for this initial mapping convolution is modality specific (reported in subsections below). The embedding dimensionality is doubled and the set of tokens is reduced by a factor of $\nu=4$ after every block. For 1D text and time series signals, the tokens are merged with a 1D convolution with kernel size 9 and stride 4. For 2D image and audio, tokens are reduced by 2D convolution with (3,3) kernel and (2,2) stride. Following~\cite{fan2021multiscale}, we also use $1,2,4,8$ heads in the MHSA of these 4 blocks. The encoder output  has $N/16$ tokens as well as a CLS embedding of size $768$. 

The proposed MAELRE architecture is tested for classification on standard and publicly available datasets of 4 different modalities: text (1D), 1D temporal sequence, audio (2D), image (2D). The datasets were selected based on 2 primary criteria: 1) number of training samples $>100$K training samples and 2) the classification tasks have immediate practical application.  The class probabilities are predicted with a single fully connected layer using the CLS embedding. 


\subsection{Text modality: MeSH prediction}\label{S:RES_MESH}
To demonstrate the performance of our model on text modality, we apply it to predict Medical Subject Headings (MeSH) used to index PubMed articles. \cite{wang2022meshup} recently published a dataset comprising 946514 full research articles in biomedicine and life sciences  along with their corresponding MeSH index terms. Because one article can be associated with multiple MeSH terms, the task we address is a multilabel classification problem.

\begin{table*}[t]
\vspace{-0.2cm}
\centering
\small
\captionsetup{font={scriptsize,stretch=0.9}} 
\caption{Performance comparison of proposed MAELRE with baselines for multilabel MeSH prediction. EBF and MiF are example and MeSH term-wise F-measures (higher is better). Both memory footprint (Mem) and FLOP were computed for a single example. The rightmost two columns depict batch size that fit in a single device with reported memory size and inference time for one sample. [Wang 2022] corresponds to~\cite{wang2022meshup} that applied a CNN-GNN model.}
\vspace{-0.3cm}
\begin{tabular}{c|c|c|c|c|c|c||c|c}
\hline
\textbf{Method} &\textbf{Token} & \textbf{EBF}$\uparrow$ &  \textbf{MiF}$\uparrow$&\textbf{Param}&\textbf{Mem}$\downarrow$&\textbf{FLOP}$\downarrow$& \textbf{bsz, GPU} & \textbf{Inf Time}$\downarrow$    \\
\hline
[Wang 2022]~\cite{wang2022meshup}& -- & 25.9 &  25.9 & -- & -- & -- & -- & --\\
\hline
Roberta~\cite{liu2019roberta} & 512 & 26.56 &  25.00 & 134M & 1233M  & 45.05G & 24, 16GB & 5.5ms\\
\hline
Longformer~\cite{beltagy2020longformer} & 2048 & 21.09 &  19.58 & 158M & 3221M  & 218.54G  & 10, 32GB & 33ms\\
\hline
BigBird~\cite{zaheer2020bigbird} & 2048 & 18.81 &  16.98 & 137M & 4010M & 183.45G & 8, 32GB & 28ms\\
\hline
OPT~\cite{zhang2022opt} & 2048 & 42.017 &  36.8 & 337M & 19523M & 782.51G & 1, 32GB & 102ms\\
\hline
Proposed-base-ELU & 2048 & 43.28 &  42.75 & 103M & 698M & 8.3G & 180, 16GB & 5.4ms\\
\hline
Proposed-base-ReLU & 2048 & 43.23 &  42.72 & 103M & 697M & 8.3G & 180, 16GB & 5.4ms\\
\hline
Proposed-base-Splus & 2048 & $\mathbf{43.35}$ &  $\mathbf{42.83}$ & 103M & 697M & 8.3G & 180, 16GB & 5.4ms\\
\hline
\hline
Proposed-base-ELU & 4096 & 44.28 &  43.77 & 103M & 746M & 16.92G & 90, 16GB & 5.5ms\\
\hline
Proposed-wide-ELU & 2048 & 53.89 & 53.94 & 205M &  1506.6M & 33.6G & 70, 16GB & 7.3ms\\
\hline
\end{tabular}
\label{T:MESH}
\vspace{-0.5cm}
\end{table*}

The average and median lengths of dataset articles are 4402 and 4102 respectively, with a standard deviation of 1691 words. Although the size of MeSH vocabulary is 28468, many MeSH tags were associated with only a few articles. To ensure that each MeSH term is learned from a statistically significant data size, we train the classifier to predict 12338 MeSH terms that were used for at least 50 articles in the dataset.  We use a common 80\%-10\%-10\% split for train, validation and test set respectively for all models.

The performance of the base proposed architecture $\{ M_1= 1, M_2 = 3, M_3 =16, M_4 = 3\}$ on this 12338 class multilabel classification is compared with classifiers constructed from a) vanilla transformer encoder, in particular RoBERTa~\cite{liu2019roberta};  b) long range encoders LongFormer~\cite{beltagy2020longformer}, BigBird~\cite{zaheer2020bigbird}; c) long range decoder OPT 350M~\cite{zhang2022opt}. The weights and code for the baseline classifiers were collected from the publicly available HuggingFace~\cite{huggingface2023transformers} repository. Each of these aforementioned classifiers were finetuned from HuggingFace pretrained models. In contrast, the proposed method was trained from scratch.

The classification target $y_j \in \{0,1\}^{C},~ C= 12338$ is binary vector where $y_j(c) = 1$ implies $c$-th MeSH term is associated with the input article $j$ and $y_j(c) = 0$ otherwise. The proposed model utilizes the tokenization scheme and token embedding of the RoBERTa model that are kept frozen during training. The RoBERTa token embeddings were projected to dimensionality $D = 96$ by 1D convolution with kernel sz 21 and stride 1. We consider both the article-wise and MeSH index-wise mean of binary cross entropy (BCE) errors between $y$ and model predictions to compute the classification cost. The same loss function consists of a weighted summation of article-wise and MeSH index-wise BCE averages of the batch were used for training.

All the models were trained with AdamW optimizer~\cite{loshchilov2019decoupled} with $\beta = [0.9, 0.999]$, weight decay $0.05$; initial learning rate $0.00005$ decayed using cosine annealing for 50 epochs with 3 warm up epochs~\cite{loshchilov2017sgdr}. Training was performed on Tesla V100 gpus with 16GB or 32GB memory with distributed data parallelism in mixed precision setting. After training, we binarize model prediction by thresholding at 0.5.  The accuracy is calculated using  measures suggested in~\cite{wang2022meshup}: 1) mean F-measure of article-wise  precision recall of MeSH terms: EBF, 2) average F-measure for MeSH index-wise precision recall of articles: MiF. Since we predict only 12338 MeSH terms (with sufficient numbers of training examples), the MeSH index F-measure MiF is computed only for these 12338 terms only. The details of accuracy, memory requirements and computational costs are summarized in Table~\ref{T:MESH}.

In addition to the accuracy values, Table~\ref{T:MESH} also reports the input token length, parameter count, maximum memory allocated during training and FLOPs required during inference for one input sample. The memory requirement and FLOPs were calculated using a pyTorch function and fvcore~\cite{fvcore} respectively. In addition, we also mention the practical implication of these two complexities via \emph{largest} batch size (bsz) that can be allocated in single device memory and inference time to predict one input on the same hardware. The table shows the accuracy of a CNN+GNN based model proposed in~\cite{wang2022meshup}. 

For the long range baselines, we utilize the maximum token length 2048 that these methods were pretrained on. Despite their efficient designs, long context models require at least 4 times more memory and 20 times more FLOPS than the proposed MAELRE for the same number of input tokens. At the same time, the proposed model achieved a higher accuracy than the long context BigBird, Longformer, OPT as well as RoBERTa. The $\ge 20\%$ higher accuracy of the proposed MAELRE over sparse attention based BigBird, Longformer supports our hypothesis that utilizing an approximate attention  among all tokens is more effective than applying original dot product attention on a subset of tokens, at least for text classification.

With 2048 tokens, the accuracy of MAELRE is $>1$\% higher and the complexity is 27 times lower for memory and 94 times lower for FLOPS than the long range OPT 350M. As the bsz of 1 and inference speed of 102ms suggest, the training and inference of OPT is significantly more costly than MAELRE. \emph{It required 12 hours to complete 1 epoch of training for OPT on a 32GB memory device as opposed to 40 mins for MAELRE on 16GB memory}. The memory cost and FLOPS of the proposed model is even lower than the 512 token RoBERTa encoder while MAELRE is 18\% more accurate than RoBERTa. The performance of MAELRE is independent of the choice of ELU as $\psi()$ function since ReLU and SoftPlus $\psi()$ function based approximate attentions achieved the same level of accuracy with the same costs.

Increasing the token length to 4096 for the proposed architecture further improved the accuracy by $\sim$ 1\% without decreasing the inference speed. \emph{We were able to train a model with 17000 input tokens with a batch size of 24 on a single 16GB memory device.} However, the MeSH prediction dataset has a mean input article length of 4402. Therefore, increasing the number of tokens beyond 4096 did not have an effect on performance. It is worth noting here that the training and inference configurations are selected for a single device and model setting and without specialized techniques. The input span can be further increased by heuristics such as subsampling words at rate $> 1$~\cite{stallone2024scaling}, device hardware specific implementations~\cite{dao2022flashattention} or by leveraging multi device distributed attention approaches~\cite{liu2023ring, li-etal-2023-sequence} as required.

We also trained a variant of our model with larger embedding dimension for MeSH classification. This variant, Proposed-wide, leverages embedding dimensions of $\{384, 384, 768, 768\}$ in the 4 computational attention blocks. The wide variant of the proposed model achieved a significant $>10$\% increase in accuracy while incurring comparable or lower memory and inference costs than the Roberta 512 token model.

\subsection{Time Series: Mosquito Species Prediction}
The proposed model was tested on the classification of 6 mosquito species from time series data. The UCR MosquitoSound dataset~\cite{Dau2019} contains the 279,566 1D sequences of mosquito wingbeats of length 3750 collected from 6 species that are evenly divided into train and test sets. The ML task is to identify the mosquito species given the wingbeat time series.

The configurations of the learning algorithm are largely same as those used for text classification except the learning rates  $1e^{-4}$ and $1e^{-5}$ used for proposed and baseline architectures respectively and different batch sizes used to fit these models in GPU. The training was performed on a single device (not distributed), in particular Tesla V100,  in a single precision setting. Furthermore, we utilized a random mix of temporal jittering, scaling, magnitude warp, time warp, random input drop available from~\cite{Iwana_2021} to augment the data and used label smoothing~\cite{szegedy2016rethinking}, MixUp~\cite{zhang2018mixup}, model exponential moving average (EMA) as learning regularizer. Apart from the batch size and learning rate, all methods were trained under the same training configuration.  
\begin{table}[h!]
\vspace{-0.0cm}
\centering
\scriptsize
\captionsetup{font={scriptsize,stretch=0.9}} 
\caption{Performance comparison for time series classification (6 class). Acc indicates top-1 accuracy. Both memory requirement (Mem) and FLOP were computed for one sample. [Cheng 2023] correspond to CNN based method of~\cite{CHENG202350}. The fvcore script threw an error and failed to compute FLOP for Reformer. }
\vspace{-0.3cm}
\begin{tabular}{c|c|c|c|c|c}
\hline
\textbf{Method} & \textbf{Token}&  \textbf{Acc}$\uparrow$ & \textbf{Param} & \textbf{Mem}$\downarrow$  & \textbf{FLOP}$\downarrow$ \\ \hline
[Cheng 2023]~\cite{CHENG202350}   & --   & 91.54   & --  & --  & --   \\ \hline
Transformer-small   & 750   & 94.22   & 21M  & 514M  & 19G   \\ \hline
Transformer-base   & 750   & 94.45   & 85M  & 1653M  & 69G   \\ \hline
Linformer~\cite{wang2020linformer}   & 3750   & 87.29   & 45M &  1886M  & 123G   \\ \hline
Performer~\cite{choromanski2021rethinking}   & 3750   & 88.39   & 39M   & 3206M  & 155G   \\ \hline
Reformer~\cite{kitaev2020reformer}   & 3712   & 86.49   &  36M  & 1827M  & --  \\ \hline
Proposed-small-ELU   & 3750   & $\mathbf{95.27}$   &  37M  & 739M  & 7G   \\ \hline
Proposed-base-ELU  & 3750  & 95.38   &  53M  & 1053M  & 10G   \\ \hline
\scriptsize{Proposed-base-ReLU}  & 3750  & 95.31   &  53M  & 1053M & 10G   \\ \hline
\scriptsize{Proposed-base-Splus}  & 3750  & $\mathbf{95.42}$   &  53M  & 1053M  & 10G   \\ \hline
\end{tabular}
\label{T:TIME}
\vspace{-0.3cm}
\end{table}
Table~\ref{T:TIME} compares the accuracy and efficiency of our architecture with the baselines. In addition, we also report the accuracy of~\cite{CHENG202350} that presents a particular variant of dilated CNN for this classification problem. The proposed MAELRE architecture outperforms all the baselines in terms of accuracy and cost for this multiclass classification of long sequence data. The base variant of MAELRE achieves a 95.42\% accuracy, which is $\sim 1\%$ higher than the next best transformer based model. The lack of capability to process long inputs restricts the context to 750 tokens for vanilla transformers, which, in turn, leads to lower accuracy. The proposed efficient base variant  utilizes $5x$ more context while requiring less than $65\%$ of memory and computation of the vanilla version. Even the smaller variant of MAELRE appears to be more accurate than the base version of transformer model. Similar to the text modality, the 3 choices of $\psi()$, ELU, ReLU, SoftPlus, were found to lead to similar accuracy for time series classification as well.

The modified attention mechanisms in long range models, Linformer, Performer, Reformer seem to be unable to match the performance of the vanilla version for the species prediction task given wingbeat sequence. These long context methods have been reported before to yield inferior accuracy than the vanilla version on various problems~\cite{CHENG202350, dao2022flashattention, phang2023investigating, zaheer2020bigbird,tay2021long}. 

\subsection{Audio: VGGSound Classification}
We apply the proposed MAELRE architecture to classify audio clips of VGGSound dataset~\cite{chen2020vggsound}. VGGSound consists of short audio clips from real life videos. We were able to collect 159223 training and 12790 evaluation clips of 309 audio categories downloaded in June 2022 from the authors of~\cite{zhu2023multiscale}. 

Following standard convention, we classify the 2D  spectrogram of audio signals into 309 classes. For this conversion, we closely follow the \emph{audio} signal processing steps of~\cite{gong2023contrastive} to generate the spectrogram. For audio, each 10-second audio waveform is first converted to a sequence of 128-dimensional log Mel filterbank (fbank) features computed with a 25ms Hanning window every 10ms. This results in a 1024(time) × 128(frequency) spectrogram. This $1024\times 128$ input matrix is converted to 8192 input tokens of size $D=96$ by a 2D convolution with stride $(4, 4)$ and kernel $(7,7)$. For clarification, we only used the audio spectrogram (filterbank) as the input for this experiment; no visual information was utilized for this classification task. The tokens were rearranged to 2D matrix before the reduction module was applied.

\begin{table}[h!]
\vspace{-0.2cm}
\centering
\scriptsize
\captionsetup{font={scriptsize,stretch=0.9}} 
\caption{Results on VGGSound classification.  Acc indicates top-1 accuracy. Both memory requirement (Mem) and FLOP were computed for one sample.}
\vspace{-0.3cm}
\begin{tabular}{c|c|c|c|c|c}
\hline
\textbf{Method} & \textbf{Token}&  \textbf{Acc}$\uparrow$ & \textbf{Param} & \textbf{Mem}$\downarrow$  & \textbf{FLOP}$\downarrow$ \\ \hline
MAST~\cite{zhu2023multiscale}  & 8192  &  57.0  & 51M  & 929M &  25G \\ \hline
AV-MAE (Audio)~\cite{gong2023contrastive}  &  512  &  57.2  & 85M  & 1157M & 45G   \\ \hline
mViTv1-small~\cite{fan2021multiscale}  &  8192  &  56.57  & 34M  & 650M  & 20G   \\ \hline
mViTv1-base~\cite{fan2021multiscale}  &  8192  &  56.74  & 50M  & 869M  & 27G   \\ \hline
mViTv2-small~\cite{li2022mvitv2}  &  8192  &  56.29  & 34M  & 649M  & 20G   \\ \hline
mViTv2-base~\cite{li2022mvitv2} &  8192  &  57.3  & 50M  & 927M  & 29G   \\ \hline
Proposed-small-ELU   &  8192  &  $\mathbf{57.47}$  & 37M  & 497M  & 16G   \\ \hline
Proposed-base-ELU   &  8192  & $\mathbf{57.85}$   & 54M  & 687M  & 23G   \\ \hline
Proposed-base-ReLU   &  8192  &  57.82  & 54M  & 687M  & 23G   \\ \hline
Proposed-base-Splus   &  8192  & 57.7   & 54M  & 708M  & 23G   \\ \hline
\end{tabular}
\label{T:AUDIO}
\vspace{-0.3cm}
\end{table}
The small and base variants of the proposed MAELRE were compared against the small and base variants of the multiscale vision transformers (mViT) v1~\cite{fan2021multiscale} and v2~\cite{li2022mvitv2}. These transformer encoder models have demonstrated impressive accuracy and efficiency in visual as well as the audio domain~\cite{zhu2023multiscale}. Both mViTv1 and mViTv2 aggregate tokens in different layers of computation with mViTv2 enhancing the representational strength using relative position embedding and residual pooling. In contrast to these models, our proposed MAELRE a) decouples token merging from attention by pooling them with 2D conv separately, b) uses different attention techniques, c) does not incorporate specialized techniques such as relative positional embedding and residual pooling. We also report the performance of the audio only AV-MAE architecture~\cite{gong2023contrastive} composed of vanilla 12 layer transformer for reference.

We followed~\cite{gong2023contrastive} to apply class balanced datasampling, frequency \& temporal masking, noise jittering for data augmentation and label smoothing, MixUp, model EMA as learning regularizer. We retain most of the training configuration utilized for text and time series classification. The training was performed with mixed precision on a single Tesla V100 32GB with the cross entropy cost function. The proposed model variants were trained with an initial learning rate of $2e^{-4}$ and a batch size of $64$ whereas those for mViT models were $1e^{-3}$ and $40$ respectively. Both proposed and mViT classifiers were finetuned from a model pretrained on image classification.


In Table~\ref{T:AUDIO}, we summarize the accuracy and costs of the proposed as well as the aforementioned baseline models. In addition, we quote the performance of the MAST encoder from~\cite{zhu2023multiscale} which also uses the mViTv2 base model for audio classification on VGGSound. Our training strategy was able to train a mViTv2 base model $0.3\%$ more accurate than MAST. The proposed MAELRE outperforms all variants of mViT and AV-MAE audio models for VGGSound classification with an accuracy of $57.85\%$. In fact, the small variant of MAELRE appears to be more accurate than the mViTv2 base model despite being significantly cheaper in terms of memory and computational costs. Although MAELRE models are larger, its efficient design leads to lower memory and computational costs than all the baselines.

All MAELRE variants with $\psi() =$ ELU, ReLU, SoftPlus consistently outperform the mViT counterparts for VGGSound classification. This implies that the specialized mechanisms such as pooling attention, relative positional embedding and residual pooling exploited in mViT models are not as beneficial in audio modality as they are in visual domain. The superior performance and better efficiency in both 1D time series and 2D audio spectrogram data (AV-MAE audio) showcase a clear advantage of the proposed model over the original transformer design.

\subsection{Vision: ImageNet Classification}
We assess our method for image classification on ImageNet1K dataset~\cite{deng2009imagenet}. ImageNet 1K consists of 1.2M training and 50K training images from 100 categories. We follow the standard convention of predicting 1000 classes by a single fully connected layer attached to the CLS embedding and train against the cross entropy loss. The performances were evaluated against the mViTv1 and v2~\cite{li2022mvitv2, fan2021multiscale} variants similar to the audio experiment. All models were trained from scratch by us using the same training recipe. 

We used the timm pytorch library~\cite{rw2019timm} for training the classifiers in a distributed and mixed precision setting.  The data augmentation, optimizer, learning regularizer and schedule configuration closely follow those suggested by MAE paper~\cite{he2022masked} and timm library scripts. Notably, we used RandAug~\cite{cubuk2020randaugment}, CutMix~\cite{yun2019cutmix}, MixUp, label smoothing, model EMA, dropout~\cite{srivastava2014dropout}, drop path~\cite{huang2016deep} with AdamW optimizer and cosine rate decay. Apart from the batch size and learning rate, that were $2e^{-3}$ for mViT and $5e^{-4}$ for MAELRE, the learning configuration was fixed for all training.

\begin{table}[h!]
\vspace{-0.2cm}
\centering
\scriptsize
\captionsetup{font={scriptsize,stretch=0.9}} 
\caption{ImageNet1K classification result.  Acc indicates top-1 accuracy. Both memory requirement (Mem) and FLOP were computed for one sample. All mViT variant results were reproduced by training from scratch.}
\vspace{-0.2cm}
\begin{tabular}{c|c|c|c|c|c}
\hline
\textbf{Method} & \textbf{Token}&  \textbf{Acc}$\uparrow$ & \textbf{Param} & \textbf{Mem}$\downarrow$  & \textbf{FLOP}$\downarrow$ \\ \hline
PoolFormerS36~\cite{yu2022metaformer}   &  3136  & 81.4 & 31M  & --  & 5G   \\ \hline
PoolFormerM36~\cite{yu2022metaformer}   &  3136  & 82.1 & 56M  & --  & 8G   \\ \hline
mViTv2-small~\cite{li2022mvitv2}   &  3136  & $\mathbf{83.23}$  & 34M  & 368M  & 6G   \\ \hline
mViTv2-base~\cite{li2022mvitv2}  &  3136  & $\mathbf{83.78}$  &  51M & 531M  & 9G   \\ \hline
mViTv1-small~\cite{fan2021multiscale}   &  3136  &  82.48 & 34M  & 360M  & 6G   \\ \hline
mViTv1-base~\cite{fan2021multiscale}   &  3136  &  82.73 & 51M  &  505M & 9G   \\ \hline
Proposed-small-ELU   &  3136  & 82.21  & 37M  & 339M& 5G   \\ \hline
Proposed-base-ELU   &  3136  & 82.97  & 54M  & 472M  & 8G   \\ \hline
\end{tabular}
\label{T:IMAGE}
\vspace{-0.3cm}
\end{table}

The results in Table~\ref{T:IMAGE} indicate that the proposed attention combination is as costly but more powerful than PoolFormer~\cite{yu2022metaformer} design that replaces attention with pooling. The accuracy of the proposed MAELRE is comparable to mViTv1 models but inferior to mViTv2 counterparts for ImageNet classification. This suggests that while the composition of approximate and dot product attention is as powerful as the pooling attention technique utilized in the mViTv1 architecture, the relative positional embedding and residual pooling, which was not used in MAELRE, are indeed beneficial for image classification. The memory and computational cost of MAELRE is lower than both mViTv1 and mViTv2 models of the same size. 

\subsection{Ablation Experiments}
\subsubsection{Attention Combination}\label{S:ABL_ATTN}
To assess the benefit of leveraging a combination of dot product attention and its approximation, we train two classifier models on text (1D) and audio (2D) modalities: 1) an architecture that applies dot product attention  and 2) another that uses only approximate attention for all layers. The base models with 2048 and 8192 tokens were tested respectively for  MeSH and audio prediction tasks.

While the variant with exclusive dot product in all layers will have a larger memory and computational footprint than MAELRE, those of all approximate variant (All-ELU-base) will depend on the token length. The costs of all approximate attention variants will become smaller than those of the proposed model with increasing token length. The comparative results in Tables~\ref{T:AB_ATTN_TXT} and~\ref{T:AB_ATTN_AUDIO} indicate: 1) using  dot product in all layers (All-dot-base) increases the complexity but does not offer considerable, if any, accuracy improvement; 2) models with approximate attention in all levels (All-ELU-base) are not as correct as MAELRE for MeSH and audio classification. In all our experiments, the proposed design of alternating attention types based on values of $N_b$ and $D_b$ was found to be more beneficial in terms of both cost and accuracy.

\begin{table}[h!]
\vspace{-0.3cm}
\centering
\scriptsize
\captionsetup{font={scriptsize,stretch=0.9}} 
\caption{Performance comparison with only dot product and only approximate~\cite{katharopoulos20rnn} attentions for MeSH prediction from text (1D).}
\vspace{-0.3cm}
\begin{tabular}{c|c|c|c|c|c}
\hline
\textbf{Method} & \textbf{EBF}$\uparrow$ &  \textbf{MiF}$\uparrow$ &\textbf{Param}&\textbf{Mem}$\downarrow$&\textbf{FLOPS}$\downarrow$    \\
\hline
Proposed-ELU &  43.28 &  42.75 & 103M & 698.53M & 8.3G \\
\hline
All-dot-base &  43.4 &  42.89 & 103M & 716.66M  & 9.28G \\
\hline
All-ELU-base &  39.8  & 39.11  & 103M & 698.92M  & 8.86G \\
\hline
\end{tabular}
\label{T:AB_ATTN_TXT}
\vspace{-0.3cm}
\end{table}


\begin{table}[h!]
\vspace{-0.25cm}
\centering
\scriptsize
\captionsetup{font={scriptsize,stretch=0.9}} 
\caption{Performance comparison with only dot product and only approximate~\cite{katharopoulos20rnn} attentions for VGGSound audio (2D) classification.}
\vspace{-0.3cm}
\begin{tabular}{c|c|c|c|c}
\hline
\textbf{Method} & \textbf{Acc}$\uparrow$ &\textbf{Param}&\textbf{Mem}$\downarrow$&\textbf{FLOPS}$\downarrow$    \\
\hline
Proposed-base-ELU  &  57.85   &  54.25M  & 687.14M  & 23.54G   \\ \hline
All-dot-base & 56.93 & 54.25M & 1347.38G  &  39.69G\\
\hline
All-ELU-base & 56.01  & 54.25M & 598.57M  & 21.08G \\
\hline
\end{tabular}
\label{T:AB_ATTN_AUDIO}
\vspace{-0.3cm}
\end{table}

\subsubsection{Token Merging}
We first analyze to what extent token merging impacts model performance and costs in different modalities. For this experiment, we use the same number of tokens in all layers while keeping all other model and training settings same as those reported in the results section. Performances are compared in terms of accuracy, memory and inference costs for text, audio and time series modalities and are reported in Table~\ref{T:AB_NO_MERGE}. The variants without token aggregation (prefix Nored-) requires $2\sim 4$ times more memory and reduces the speed by an order of magnitude. Despite the increase in computational cost, these variants were unable to match the accuracy of the proposed design in audio and time series modalities. Retaining the same number of tokens appears to improve the accuracy of text classification on MeSH data by $>2$\% over our model. However, as we report in Table~\ref{T:MESH},  the proposed method can achieve a $\sim 10$\% higher accuracy at comparable memory but lower latency when we increase the embedding dimension (Proposed-wide-ELU).
\begin{table}[h!]
\vspace{-0.25cm}
\centering
\scriptsize
\captionsetup{font={scriptsize,stretch=0.9}} 
\caption{Performance comparison with (Proposed-ELU) and without token aggregation (Nored-ELU) on text, audio and time series data.}
\vspace{-0.3cm}
\begin{tabular}{c|c|c|c|c|c}
\hline
\textbf{Method} & \textbf{Mod} &\textbf{Acc}$\uparrow$ &\textbf{Param}&\textbf{Mem}$\downarrow$&\textbf{FLOPS}$\downarrow$    \\
\hline
Proposed-ELU & Text & (43.28,  42.75) & 103M & 698M & 8.3G \\ 
\hline
Nored-ELU  & Text & (45.47, 44.94)   &  101.3M  & 1434M  & 104G   \\ 
\hline
\hline
Proposed-ELU  & Series & 95.38   &  53M  & 1053.5M  & 10.06G   \\ \hline
Nored-ELU  & Series &  89.65  &  52.5M  & 3387.85M  & 405.86G   \\ 
\hline
\hline
Proposed-ELU  & Audio & 57.85   &  54.25M  & 687.14M  & 23.54G   \\ \hline
Nored-ELU  & Audio &  52.67  &  51.84M  & 2821.91M  & 743.23G   \\ 
\hline
\end{tabular}
\label{T:AB_NO_MERGE}
\vspace{-0.3cm}
\end{table}

Next, we investigate the particular technique that we utilized for token aggregation. To validate our choice of token merging using convolution with a stride $\tau$ = 4 (or [2,2] for 2D data), we compare the performances of the proposed MAELRE with variants that aggregate tokens by 1) average pooling with kernel size 9 ([3,3] for 2D data)~\cite{li2022efficientformer,yu2022metaformer} , and 2) linear combination that maps $N_b$ tokens to $N_b/4$ by a single perception layer. 
\begin{table}[h!]
\vspace{-0.3cm}
\centering
\scriptsize
\captionsetup{font={scriptsize,stretch=0.9}} 
\caption{Performance comparison of different token aggregation methods, specifically pooling and linear projection, on on time series (1D) and audio (2D) data.}
\vspace{-0.3cm}
\begin{tabular}{c|c|c|c|c|c}
\hline
\textbf{Method} & \textbf{Mod} &\textbf{Acc}$\uparrow$ &\textbf{Param}&\textbf{Mem}$\downarrow$&\textbf{FLOPS}$\downarrow$    \\
\hline
Proposed-ELU  & Series & 95.38   &  53M  & 1053.5M  & 10.06G   \\ \hline
Merge-pooling & Series & 93.92 & 51.84M  & 1020.68M  &  39.37G\\
\hline
Merge-linear & Series & 88.92 & 55.59M  & 1094.55M  &  40.59G\\
\hline
\hline
Proposed-ELU  & Audio & 57.85   &  54.25M  & 687.14M  & 23.54G   \\ \hline
Merge-pooling & Audio & 54.59 & 52.5M &  676.04 & 23.06G \\
\hline
Merge-linear & Audio & 49.06  & 70.4M & 775.82M  &  24.78G\\
\hline
\end{tabular}
\label{T:AB_MRG_TIME}
\vspace{-0.4cm}
\end{table}


The inferior performance of pooling  (Merge-pooling) in Table~\ref{T:AB_MRG_TIME} implies that local averaging is not as effective as convolution to aggregate necessary information in the proposed architecture. A reduction through linear transformation between $N_b$ and $N_b/4$ tokens may intuitively appear to be aligned with the all to all attention methodology and therefore anticipated to perform optimal reduction. However, the performance of Merge-linear in Table~\ref{T:AB_MRG_TIME} suggests that it may be difficult to learn the weights of such transformation in practice and may even be redundant. For many problems, a local consolidation of tokens may be sufficient for long context processing, which can be effectively learned by convolution.

\section{Conclusion}
This paper introduces an efficient encoder architecture, MAELRE, for long range problems. The proposed design intelligently combines token merging with approximate attention methods to amplify efficiency gained from these techniques while maintaining accuracy. The proposed MAELRE attained higher classification accuracy than existing efficient and long range methods while incurring significantly smaller memory and computational costs in 4 different modalities: text, audio, time series and vision. The proposed efficient architecture is designed for single device and model setting -- applying multidevice distributed techniques could potentially magnify the input context multiple folds.

{\footnotesize
\bibliography{MAELRE}

\begin{thebibliography}{10}\itemsep=-1pt

\bibitem{ainslie2023groupquery}
Joshua Ainslie, James Lee-Thorp, Michiel de Jong, Yury Zemlyanskiy, Federico Lebron, and Sumit Sanghai.
\newblock {GQA}: Training generalized multi-query transformer models from multi-head checkpoints.
\newblock In {\em The 2023 Conference on Empirical Methods in Natural Language Processing}, 2023.

\bibitem{alayrac2022flamingo}
Jean-Baptiste Alayrac, Ana Recasens, Antoni Miech, Marjan Ghazvininejad, Marc'Aurelio Ranzato, Yuval Avidor, Yuandong Zhai, Minseok Lee, Naman Goyal, Wei-Cheng Hsu, et~al.
\newblock Flamingo: a visual language model for few-shot learning.
\newblock In {\em Advances in Neural Information Processing Systems (NeurIPS)}, 2022.

\bibitem{beltagy2020longformer}
Iz Beltagy, Matthew~E Peters, and Arman Cohan.
\newblock Longformer: The long-document transformer.
\newblock {\em arXiv preprint arXiv:2004.05150}, 2020.

\bibitem{bolya2022token}
Daniel Bolya, Cheng-Yang Fu, Xiaoliang Dai, Peizhao Zhang, Christoph Feichtenhofer, and Judy Hoffman.
\newblock Token merging: Your vit but faster.
\newblock {\em arXiv preprint arXiv:2210.09461}, 2022.

\bibitem{bulatov2023scaling}
Aydar Bulatov, Yuri Kuratov, Yermek Kapushev, and Mikhail~S. Burtsev.
\newblock Scaling transformer to 1m tokens and beyond with rmt.
\newblock {\em arXiv preprint arXiv:2304.11062}, 2023.

\bibitem{chen2020vggsound}
Honglie Chen, Weidi Xie, Andrea Vedaldi, and Andrew Zisserman.
\newblock Vggsound: A large-scale audio-visual dataset.
\newblock In {\em Proceedings of the IEEE International Conference on Acoustics, Speech, and Signal Processing (ICASSP)}, pages 721--725. IEEE, 2020.

\bibitem{CHENG202350}
Lei Cheng, Ruslan Khalitov, Tong Yu, Jing Zhang, and Zhirong Yang.
\newblock Classification of long sequential data using circular dilated convolutional neural networks.
\newblock {\em Neurocomputing}, 518:50--59, 2023.

\bibitem{child2019generating}
Rewon Child, Scott Gray, Alec Radford, and Ilya Sutskever.
\newblock Generating long sequences with sparse transformers.
\newblock {\em arXiv preprint arXiv:1904.10509}, 2019.

\bibitem{choromanski2021rethinking}
Krzysztof Choromanski, Valerii Likhosherstov, David Dohan, Xingyou Song, Alex Gane, Tamas Sarlos, Peter Hawkins, Jared Davis, Afroz Mohiuddin, Lukasz Kaiser, David Belanger, Lucy Colwell, and Adrian Weller.
\newblock Rethinking attention with performers.
\newblock In {\em International Conference on Learning Representations}, 2021.

\bibitem{cubuk2020randaugment}
Ekin~D Cubuk, Barret Zoph, Jonathon Shlens, and Quoc~V Le.
\newblock Randaugment: Practical automated data augmentation with a reduced search space.
\newblock In {\em Advances in Neural Information Processing Systems (NeurIPS) Workshops}, 2020.

\bibitem{dai2020funnel}
Zihang Dai, Guokun Lai, Yiming Yang, and Quoc~V. Le.
\newblock Funnel-transformer: Filtering out sequential redundancy for efficient language processing.
\newblock In {\em Advances in Neural Information Processing Systems (NeurIPS)}, 2020.

\bibitem{dao2023flashattention2}
Tri Dao.
\newblock Flash{A}ttention-2: Faster attention with better parallelism and work partitioning.
\newblock In {\em International Conference on Learning Representations (ICLR)}, 2024.

\bibitem{dao2022flashattention}
Tri Dao, Daniel~Y. Fu, Stefano Ermon, Atri Rudra, and Christopher R{\'e}.
\newblock Flashattention: Fast and memory-efficient exact attention with io-awareness.
\newblock In {\em Advances in Neural Information Processing Systems (NeurIPS)}, volume~35, pages 16344--16359, 2022.

\bibitem{Dau2019}
Hoang~Anh Dau, Anthony Bagnall, Kaveh Kamgar, Chin-Chia~Michael Yeh, Yan Zhu, Shaghayegh Gharghabi, Chotirat~Ann Ratanamahatana, and Eamonn Keogh.
\newblock The ucr time series classification archive, 2019.
\newblock Accessed: 2024-12-29.

\bibitem{deepseekai2024deepseekv2}
DeepSeek-AI.
\newblock Deepseek-v2: A strong, economical, and efficient mixture-of-experts language model.
\newblock {\em arXiv}, 2405.04434, 2024.

\bibitem{deng2009imagenet}
Jia Deng, Wei Dong, Richard Socher, Li-Jia Li, Kai Li, and Li Fei-Fei.
\newblock Imagenet: A large-scale hierarchical image database.
\newblock In {\em 2009 IEEE Conference on Computer Vision and Pattern Recognition}, pages 248--255. IEEE, 2009.

\bibitem{fvcore}
FacebookAI.
\newblock fvcore: A collection of core libraries for computer vision.
\newblock \url{https://github.com/facebookresearch/fvcore}, 2019.
\newblock Accessed: 2024-12-27.

\bibitem{fan2021multiscale}
Haoqi Fan, Bo Xiong, Karttikeya Mangalam, Yanghao Li, Zhicheng Yan, Jitendra Malik, and Christoph Feichtenhofer.
\newblock Multiscale vision transformers.
\newblock In {\em Proceedings of the IEEE/CVF International Conference on Computer Vision (ICCV)}, pages 6824--6835, 2021.

\bibitem{fu2022hungry}
Daniel~Y. Fu, Tri Dao, Khaled~K. Saab, Armin~W. Thomas, Atri Rudra, and Christopher R{\'e}.
\newblock Hungry hungry hippos: Towards language modeling with state space models.
\newblock {\em arXiv preprint arXiv:2212.14052}, 2022.

\bibitem{geva2023dissecting}
Mor Geva, Jasmijn Bastings, Katja Filippova, and Amir Globerson.
\newblock Dissecting recall of factual associations in auto-regressive language models.
\newblock {\em arXiv preprint arXiv:2304.14767}, 2023.

\bibitem{gong2023contrastive}
Yuan Gong, Andrew Rouditchenko, Alexander~H. Liu, David Harwath, Leonid Karlinsky, Hilde Kuehne, and James~R. Glass.
\newblock Contrastive audio-visual masked autoencoder.
\newblock In {\em The Eleventh International Conference on Learning Representations}, 2023.

\bibitem{goyal2020powerbert}
Saurabh Goyal, Anamitra~R. Choudhury, Saurabh~M. Raje, Venkatesan~T. Chakaravarthy, Yogish Sabharwal, and Ashish Verma.
\newblock Power-bert: Accelerating bert inference via progressive word-vector elimination.
\newblock In {\em Proceedings of the 37th International Conference on Machine Learning (ICML)}, 2020.

\bibitem{graham2021levit}
Benjamin Graham, Alaaeldin El-Nouby, Hugo Touvron, Pierre Stock, Armand Joulin, Herv{\'e} J{\'e}gou, and Matthijs Douze.
\newblock Levit: a vision transformer in convnet’s clothing for faster inference.
\newblock In {\em Proceedings of the IEEE/CVF International Conference on Computer Vision (ICCV)}, pages 12259--12269, 2021.

\bibitem{gu2023mamba}
Albert Gu and Tri Dao.
\newblock Mamba: Linear-time sequence modeling with selective state spaces.
\newblock {\em arXiv preprint arXiv:2312.00752}, 2023.

\bibitem{guo2022longt5}
Mandy Guo, Joshua Ainslie, David Uthus, and Santiago Onta{\~n}{\'o}n.
\newblock Longt5: Efficient text-to-text transformer for long sequences.
\newblock In {\em Proceedings of the 2022 Conference of the North American Chapter of the Association for Computational Linguistics: Human Language Technologies}, pages 7247--7259, 2022.

\bibitem{han2023flatten}
Dongchen Han, Xuran Pan, Yizeng Han, Shiji Song, and Gao Huang.
\newblock Flatten transformer: Vision transformer using focused linear attention.
\newblock In {\em Proceedings of the IEEE/CVF International Conference on Computer Vision (ICCV)}, 2023.

\bibitem{he2022masked}
Kaiming He, Xiangyu Chen, Saining Xie, Yanghao Li, Piotr Doll{\'a}r, and Ross Girshick.
\newblock Masked autoencoders are scalable vision learners.
\newblock {\em Proceedings of the IEEE/CVF Conference on Computer Vision and Pattern Recognition}, pages 16000--16009, 2022.

\bibitem{heo2021rethinking}
Byeongho Heo, Sangdoo Yun, Dongyoon Han, Sanghyuk Chun, Junsuk Choe, and Seong~Joon Oh.
\newblock Rethinking spatial dimensions of vision transformers.
\newblock In {\em Proceedings of the IEEE/CVF International Conference on Computer Vision (ICCV)}, pages 11936--11945, 2021.

\bibitem{hofmann2008kernel}
Thomas Hofmann, Bernhard Schölkopf, and Alexander~J. Smola.
\newblock Kernel methods in machine learning.
\newblock {\em The MIT Press}, 2008.

\bibitem{huang2016deep}
Gao Huang, Yu Sun, Zhuang Liu, Daniel Sedra, and Kilian~Q Weinberger.
\newblock Deep networks with stochastic depth.
\newblock In {\em European Conference on Computer Vision (ECCV)}, pages 646--661. Springer, 2016.

\bibitem{huggingface2023transformers}
HuggingFace.
\newblock Transformers: State-of-the-art machine learning for pytorch, tensorflow, and jax.
\newblock \url{https://github.com/huggingface/transformers}, 2023.

\bibitem{Iwana_2021}
Brian~Kenji Iwana and Seiichi Uchida.
\newblock An empirical survey of data augmentation for time series classification with neural networks.
\newblock {\em {PLOS} {ONE}}, 16(7):e0254841, jul 2021.

\bibitem{katharopoulos20rnn}
A. Katharopoulos, A. Vyas, N. Pappas, and F. Fleuret.
\newblock Transformers are rnns: Fast autoregressive transformers with linear attention.
\newblock In {\em Proceedings of the International Conference on Machine Learning (ICML)}, 2020.

\bibitem{kitaev2020reformer}
Nikita Kitaev, Lukasz Kaiser, and Anselm Levskaya.
\newblock Reformer: The efficient transformer.
\newblock In {\em International Conference on Learning Representations}, 2020.

\bibitem{li-etal-2023-sequence}
Shenggui Li, Fuzhao Xue, Chaitanya Baranwal, Yongbin Li, and Yang You.
\newblock Sequence parallelism: Long sequence training from system perspective.
\newblock In {\em Proceedings of the 61st Annual Meeting of the Association for Computational Linguistics (Volume 1: Long Papers)}, pages 2391--2404, Toronto, Canada, 2023. Association for Computational Linguistics.

\bibitem{li2022mvitv2}
Yanghao Li, Chao-Yuan Wu, Haoqi Fan, Karttikeya Mangalam, Bo Xiong, Jitendra Malik, and Christoph Feichtenhofer.
\newblock Mvitv2: Improved multiscale vision transformers for classification and detection.
\newblock In {\em Proceedings of the IEEE/CVF Conference on Computer Vision and Pattern Recognition (CVPR)}, pages 4804--4814, 2022.

\bibitem{li2022efficientformer}
Yanyu Li, Geng Yuan, Yang Wen, Ju Hu, Georgios Evangelidis, Sergey Tulyakov, Yanzhi Wang, and Jian Ren.
\newblock Efficientformer: Vision transformers at mobilenet speed.
\newblock In {\em Advances in Neural Information Processing Systems (NeurIPS)}, 2022.

\bibitem{liu2023ring}
Hao Liu, Matei Zaharia, and Pieter Abbeel.
\newblock Ring attention with blockwise transformers for near-infinite context.
\newblock {\em arXiv preprint arXiv:2310.01889}, 2023.

\bibitem{liu2019roberta}
Yinhan Liu, Myle Ott, Naman Goyal, Jingfei Du, Mandar Joshi, Danqi Chen, Omer Levy, Mike Lewis, Luke Zettlemoyer, and Veselin Stoyanov.
\newblock Roberta: A robustly optimized bert pretraining approach.
\newblock {\em arXiv preprint arXiv:1907.11692}, 2019.

\bibitem{liu2021swin}
Ze Liu, Yutong Lin, Yue Cao, Han Hu, Yixuan Wei, Zheng Zhang, Stephen Lin, and Baining Guo.
\newblock Swin transformer: Hierarchical vision transformer using shifted windows.
\newblock In {\em Proceedings of the IEEE/CVF International Conference on Computer Vision (ICCV)}, pages 10012--10022, 2021.

\bibitem{loshchilov2017sgdr}
Ilya Loshchilov and Frank Hutter.
\newblock Sgdr: Stochastic gradient descent with warm restarts.
\newblock In {\em International Conference on Learning Representations}, 2017.

\bibitem{loshchilov2019decoupled}
Ilya Loshchilov and Frank Hutter.
\newblock Decoupled weight decay regularization.
\newblock In {\em International Conference on Learning Representations}, 2019.

\bibitem{marin2023token}
Dmitrii Marin, Jen-Hao~Rick Chang, Anurag Ranjan, Anish Prabhu, Mohammad Rastegari, and Oncel Tuzel.
\newblock Token pooling in vision transformers for image classification.
\newblock In {\em Proceedings of the IEEE/CVF Winter Conference on Applications of Computer Vision (WACV)}, 2023.

\bibitem{meng2022locating}
Kevin Meng, David Bau, Alex Andonian, and Yonatan Belinkov.
\newblock Locating and editing factual associations in gpt.
\newblock In {\em Advances in Neural Information Processing Systems}, volume~35, pages 17204--17219, 2022.

\bibitem{Pan_2021_ICCV}
Zizheng Pan, Bohan Zhuang, Jing Liu, Haoyu He, and Jianfei Cai.
\newblock Scalable vision transformers with hierarchical pooling.
\newblock In {\em Proceedings of the IEEE/CVF International Conference on Computer Vision (ICCV)}, pages 377--386, October 2021.

\bibitem{phang2023investigating}
Jason Phang, Yao Zhao, and Peter Liu.
\newblock Investigating efficiently extending transformers for long input summarization.
\newblock In {\em Proceedings of the 2023 Conference on Empirical Methods in Natural Language Processing}, Singapore, 2023.

\bibitem{rabe2021selfattention}
Markus~N. Rabe and Charles Staats.
\newblock Self-attention does not need $o(n^2)$ memory.
\newblock {\em arXiv preprint arXiv:2112.05682}, 2021.

\bibitem{radford2021learning}
Alec Radford, Jong~Wook Kim, Chris Hallacy, Aditya Ramesh, Gabriel Goh, Sandhini Agarwal, Girish Sastry, Amanda Askell, Pamela Mishkin, Jack Clark, Gretchen Krueger, and Ilya Sutskever.
\newblock Learning transferable visual models from natural language supervision.
\newblock {\em arXiv preprint arXiv:2103.00020}, 2021.

\bibitem{shazeer19multiquery}
Noam Shazeer.
\newblock Fast transformer decoding: One write-head is all you need.
\newblock {\em CoRR}, abs/1911.02150, 2019.

\bibitem{srivastava2014dropout}
Nitish Srivastava, Geoffrey Hinton, Alex Krizhevsky, Ilya Sutskever, and Ruslan Salakhutdinov.
\newblock Dropout: A simple way to prevent neural networks from overfitting.
\newblock {\em Journal of Machine Learning Research}, 15(56):1929--1958, 2014.

\bibitem{stallone2024scaling}
Matt Stallone, Vaibhav Saxena, Leonid Karlinsky, Bridget McGinn, Tim Bula, Mayank Mishra, Adriana Meza~Soria, Gaoyuan Zhang, Aditya Prasad, Yikang Shen, Saptha Surendran, Shanmukha Guttula, Hima Patel, Parameswaran Selvam, Xuan-Hong Dang, Yan Koyfman, Atin Sood, Rogerio Feris, Nirmit Desai, David~D. Cox, Ruchir Puri, and Rameswar Panda.
\newblock Scaling granite code models to 128k context.
\newblock {\em arXiv preprint arXiv:2407.13739}, 2024.

\bibitem{szegedy2016rethinking}
Christian Szegedy, Vincent Vanhoucke, Sergey Ioffe, Jon Shlens, and Zbigniew Wojna.
\newblock Rethinking the inception architecture for computer vision.
\newblock In {\em Proceedings of the IEEE Conference on Computer Vision and Pattern Recognition (CVPR)}, pages 2818--2826, 2016.

\bibitem{tay2020sparse}
Yi Tay, Dara Bahri, Liu Yang, Donald Metzler, and Da-Cheng Juan.
\newblock Sparse sinkhorn attention.
\newblock In {\em Proceedings of the 37th International Conference on Machine Learning}, pages 9438--9447, 2020.

\bibitem{tay2021long}
Yi Tay, Mostafa Dehghani, Dara Bahri, and Donald Metzler.
\newblock Long range arena: A benchmark for efficient transformers.
\newblock In {\em International Conference on Learning Representations (ICLR)}, 2021.

\bibitem{vaswani2017attention}
Ashish Vaswani, Noam Shazeer, Niki Parmar, Jakob Uszkoreit, Llion Jones, Aidan~N Gomez, Lukasz Kaiser, and Illia Polosukhin.
\newblock Attention is all you need.
\newblock In {\em Advances in Neural Information Processing Systems}, volume~30. Curran Associates, Inc., 2017.

\bibitem{wang2022interpretability}
Kevin Wang, Alexandre Variengien, Arthur Conmy, Buck Shlegeris, and Jacob Steinhardt.
\newblock Interpretability in the wild: a circuit for indirect object identification in gpt-2 small.
\newblock {\em arXiv preprint arXiv:2211.00593}, 2022.

\bibitem{wang2020linformer}
Sinong Wang, Belinda~Z. Huang, Krzysztof Choromanski, Valerii Likhosherstov, and Adrian Weller.
\newblock Linformer: Self-attention with linear complexity.
\newblock In {\em Advances in Neural Information Processing Systems}, volume~33, pages 5290--5300. Curran Associates, Inc., 2020.

\bibitem{wang2022meshup}
Xindi Wang, Robert~E Mercer, and Frank Rudzicz.
\newblock Meshup: A corpus for full text biomedical document indexing.
\newblock {\em arXiv preprint arXiv:2204.13604}, 2022.

\bibitem{rw2019timm}
Ross Wightman.
\newblock Pytorch image models.
\newblock \url{https://github.com/rwightman/pytorch-image-models}, 2019.

\bibitem{xiong2024effective}
Wenhan Xiong, Jingyu Liu, Igor Molybog, Hejia Zhang, Prajjwal Bhargava, Rui Hou, Louis Martin, Rashi Rungta, Karthik~Abinav Sankararaman, Barlas Oguz, Madian Khabsa, Han Fang, Yashar Mehdad, Sharan Narang, Kshitiz Malik, Angela Fan, Shruti Bhosale, Sergey Edunov, Mike Lewis, Sinong Wang, and Hao Ma.
\newblock Effective long-context scaling of foundation models.
\newblock In {\em Proceedings of the 2024 Conference of the North American Chapter of the Association for Computational Linguistics: Human Language Technologies}, 2024.

\bibitem{yin2022avit}
Hongxu Yin, Arash Vahdat, Jose Alvarez, Arun Mallya, Jan Kautz, and Pavlo Molchanov.
\newblock A-vit: Adaptive tokens for efficient vision transformer.
\newblock In {\em Proceedings of the IEEE/CVF Conference on Computer Vision and Pattern Recognition (CVPR)}, pages 10856--10865, 2022.

\bibitem{yu2023megabyte}
Lili Yu, D{\'a}niel Simig, Colin Flaherty, Armen Aghajanyan, Luke Zettlemoyer, and Mike Lewis.
\newblock Megabyte: Predicting million-byte sequences with multiscale transformers.
\newblock {\em arXiv preprint arXiv:2305.07185}, 2023.

\bibitem{yu2022metaformer}
Weihao Yu, Cheng Si, Yujun Yu, Minghao Luo, Tao Zhou, and Jiashi Du.
\newblock Metaformer is actually what you need for vision.
\newblock {\em Proceedings of the IEEE/CVF Conference on Computer Vision and Pattern Recognition (CVPR)}, pages 10819--10829, 2022.

\bibitem{yun2019cutmix}
Sangdoo Yun, Dongyoon Han, Seong~Joon Oh, Sanghyuk Chun, Junsuk Choe, and Youngjoon Yoo.
\newblock Cutmix: Regularization strategy to train strong classifiers with localizable features.
\newblock In {\em Proceedings of the IEEE International Conference on Computer Vision (ICCV)}, pages 6023--6032, 2019.

\bibitem{zaheer2020bigbird}
Manzil Zaheer, Guru Guruganesh, Kumar~Avinava Dubey, Joshua Ainslie, Chris Alberti, Santiago Ontanon, Philip Pham, Anirudh Ravula, Qifan Wang, Li Yang, and Amr Ahmed.
\newblock Big bird: Transformers for longer sequences.
\newblock In {\em Advances in Neural Information Processing Systems (NeurIPS)}, 2020.

\bibitem{zhang2018mixup}
Hongyi Zhang, Moustapha Cisse, Yann~N Dauphin, and David Lopez-Paz.
\newblock mixup: Beyond empirical risk minimization.
\newblock In {\em International Conference on Learning Representations}, 2018.

\bibitem{zhang2022opt}
Susan Zhang, Stephen Roller, Naman Goyal, Mikel Artetxe, Moya Chen, Shuohui Chen, Christopher Dewan, Mona Diab, Xian Li, Xi~Victoria Lin, et~al.
\newblock Opt: Open pre-trained transformer language models.
\newblock {\em arXiv preprint arXiv:2205.01068}, 2022.

\bibitem{zhao2022adavit}
Zhize Zhao, Hong Xu, Bo Zhang, Haoyi Zhang, Zhe Wang, Xiaohui Zhai, Lijun Zhang, and Zhiqiang Wang.
\newblock Adavit: Adaptive vision transformers for efficient image recognition.
\newblock In {\em Proceedings of the IEEE/CVF Conference on Computer Vision and Pattern Recognition (CVPR)}, pages 10650--10659, 2022.

\bibitem{zhu2021longshort}
Chen Zhu, Wei Ping, Chaowei Xiao, Mohammad Shoeybi, Tom Goldstein, Anima Anandkumar, and Bryan Catanzaro.
\newblock Long-short transformer: Efficient transformers for language and vision.
\newblock In {\em Advances in Neural Information Processing Systems}, volume~34, pages 16191--16203, 2021.

\bibitem{zhu2023multiscale}
Wentao Zhu and Mohamed Omar.
\newblock Multiscale audio spectrogram transformer for efficient audio classification.
\newblock {\em arXiv preprint arXiv:2303.10757}, 2023.

\end{thebibliography}
\bibliographystyle{ieee_fullname}}

\end{document}